\documentclass{INTERSPEECH2023}
\usepackage{cite}
\usepackage{marvosym}

\interspeechcameraready

\title{Emotional Talking Head Generation based on Memory-Sharing and Attention-Augmented Networks}

\name{Jianrong Wang$^{1}$ \qquad Yaxin Zhao$^{2}$ \qquad Li Liu$^{3}$$^{(\textrm{\Letter})}$ \thanks{This work was supported by the National Natural Science Foundation of China (No. 61977049) and the National Natural Science Foundation of
China (No. 62101351)}\qquad Tianyi Xu$^{1}$ \qquad Qi Li$^{4}$ \qquad Sen Li$^{1}$}

\address{
  $^{1}$College of Intelligence and Computing, Tianjin University, Tianjin, China\\
  $^{2}$Tianjin International Engineering Institute, Tianjin University, Tianjin, China\\
  $^{3}$The Hong Kong University of Science and Technology (Guangzhou), Guangzhou, China\\
  $^{4}$School of Electrical and Information Engineering, Tianjin University, Tianjin, China
  }

\email{ avrillliu@hkust-gz.edu.cn}

\begin{document}

\maketitle
 
\begin{abstract}
Given an audio clip and a reference face image, the goal of the talking head generation is to generate a high-fidelity talking head video. Although some audio-driven methods of generating talking head videos have made some achievements in the past, most of them only focused on lip and audio synchronization and lack the ability to reproduce the facial expressions of the target person. To this end, we propose a talking head generation model consisting of a \textbf{M}emory-\textbf{S}haring \textbf{E}motion \textbf{F}eature extractor (MSEF) and an \textbf{A}ttention-\textbf{A}ugmented \textbf{T}ranslator based on \textbf{U}-net (AATU). Firstly, MSEF can extract implicit emotional auxiliary features from audio to estimate more accurate emotional face landmarks.~Secondly, AATU acts as a translator between the estimated landmarks and the photo-realistic video frames. Extensive qualitative and quantitative experiments have shown the superiority of the proposed method to the previous works. Codes will be made publicly available.

\end{abstract}

\vspace{-0.2cm}
\section{Introduction}

Audio-driven realistic talking head video generation plays a very important role in multiple applications, such as film making \cite{kim2019neural}, video bandwidth reduction \cite{wang2021one}, virtual avatars animation \cite{lu2021live,wang2021cross} and video conference \cite{guo2021ad}, etc. According to the previous work \cite{ji2022eamm,  
 zhou2021pose}, an ideal realistic talking head video should satisfy the following requirements, \emph{i.e.}, (1) the identity needs to be consistent with the target person, (2) the lip movements need to be synchronized with the audio content, (3) the videos should have natural facial expressions and head movements.

In the literature, some previous works have focused on generating lip-synchronized talking head videos \cite{sadoughi2019speech,prajwal2020lip,wang2022residual}, but they ignored the facial expressions modeling. In recent years, there has been some work on generating expression-controlled talking head videos. Blinking motions were added in \cite{vougioukas2020realistic,sinha2020identity} to improve the realism by synthesizing talking head videos, but the results were still unsatisfactory, \emph{i.e.}, the facial muscles were stiff. \cite{ji2021audio} relied on neutral video recordings of the target person to generate
emotional talking head videos, but the facial expressiveness of the generated results was still insufficient. \cite{ji2022eamm} designed a model to generate a talking head video that is emotionally consistent with an emotional source video by accepting four inputs, namely an identity reference image, an audio clip, a predefined pose video and the emotional source video. However, the video-driven based approach is limited by bandwidth, storage space, etc., and is not applicable in some application cases, such as bandwidth-constrained video conferencing.

Based on the above research and analysis, our work is to design an audio-driven talking head generation model that accepts two inputs, \emph{i.e.}, an emotional audio clip and a reference facial image with the same emotion. The outputs are highly realistic videos of the target person. We believe that with the rapid development of photographic devices such as mobile phones and cameras, it should be easy to obtain such inputs. However, there are still two challenges in implementing such a model. Firstly, the facial pose of a person varies greatly across different emotional states. Secondly, rich facial expressions produce complex skin textures and facial shadows. To make the generated emotional talking head videos more realistic, we not only need to accurately predict the emotional facial landmarks, but also need to render the facial details during the regression from the landmarks to the images.

To solve these problems, we propose a new two-staged emotional talking head generation model. More precisely, in the first stage, because the emotional information in the audio is closely related to facial expressions, we explicitly extract the emotional features hidden in the audio as the auxiliary information. We train a \emph{Memory-Sharing Emotional Feature extractor} (MSEF) in a supervised way, and propose a joint loss to change the optimization direction of the model to further improve the accuracy of the predicted landmarks. MSEF implicitly takes into account the relationship between different samples through the memory-sharing module with linear complexity, which is of great significance for extracting emotional features in audio. In the second stage, the predicted landmarks and the reference face image are fed into the \emph{Attention-Augmented Translator} based on \emph{U-net} (AATU) to generate photo-realistic talking head videos. AATU aims to focus on shallow details and important semantic features of the network simultaneously, reducing the loss of useful information and improving the model's performance, so that the output image can maintain more details such as skin texture and facial shadows of the target person.

\begin{figure*}[htbp]
\centerline{\includegraphics[width=0.9\linewidth]{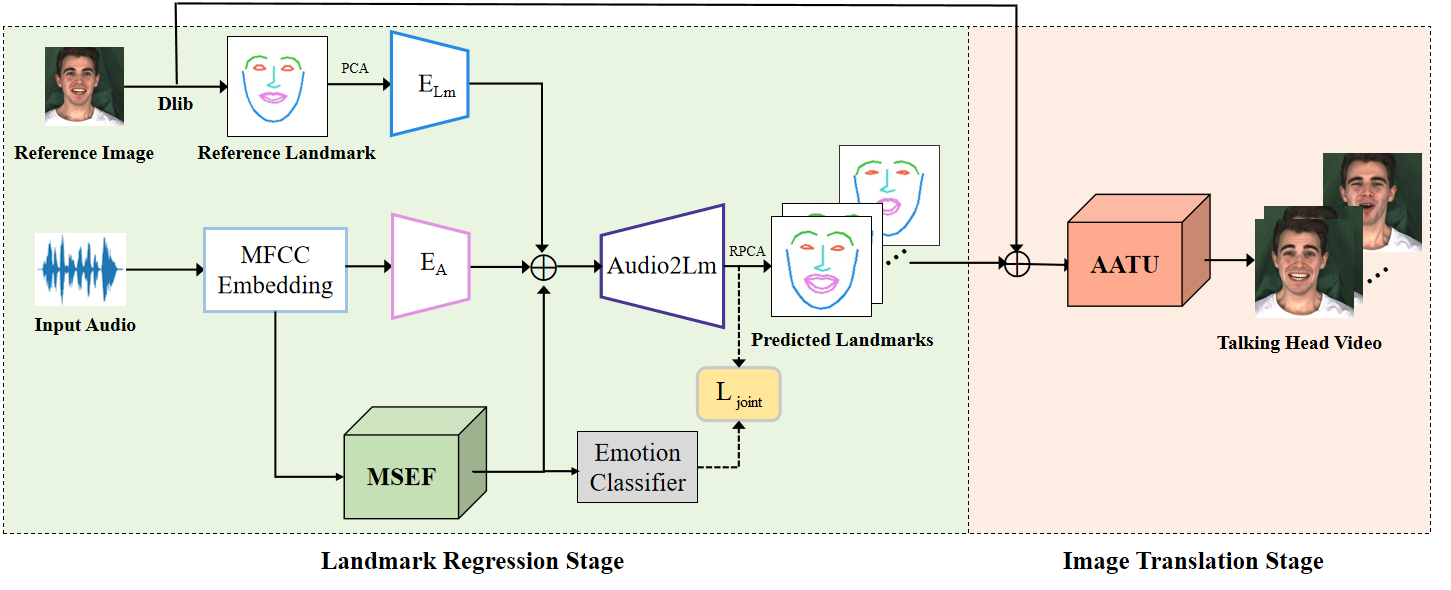}}
\caption{Overview of our proposed talking head generation network. First of all, we extract implied emotional features from audio signals through MSEF module. $E_{Lm}$ and $E_{A}$ are landmark encoder and MFCC encoder, respectively. Secondly, we concatenate the landmark features, MFCC features and emotional features, and feed them into the $Audio2Lm$ module to predict dynamic face landmarks. Finally, we splice the predicted landmarks with the reference face image by channel, and render them into photo-realistic talking head video frames through the AATU module.}
\label{fig1}
\end{figure*}

To sum up, our contributions can be summarized as follows.

\begin{itemize}
\item We propose a novel model, a memory-sharing emotional feature extractor, to extract emotional features from audio signals. Using the extracted auxiliary features, the network can predict emotional face landmarks more accurately than previous works.

\item An attention-augmented translator based on U-net is proposed to generate photo-realistic and emotional talking head video frames, \emph{e.g.}, skin texture and
facial shadows.

\item Qualitative and quantitative experiments on the MEAD dataset show that the model achieves high-quality emotional talking head video generation, which is significantly superior to previous works.

\end{itemize}

\vspace{-0.2cm}
\section{Releted Work}

In the literature, there are two general approaches to talking head video generation. One is the end-to-end mapping from audio to talking head video \cite{liang2022expressive, ye2022audio,wang2023memory}, and the other is the generation of talking head video through intermediate features, such as landmarks \cite{das2020speech, zhou2020makelttalk,liu2020re} and 3DMM parameters \cite{zhang2021facial, richard2021audio}. 

The large number of parameters in the end-to-end talking head generation model often leads to the model overfitting the training data. Furthermore, the
networks tend to be more concerned with lip and audio synchronization, making it difficult to generate face animations of various natural facial poses and expressions. To this end, Zhou \emph{et al.} \cite{chen2019hierarchical} first proposed a cascading model, \emph{i.e.}, first mapping audio to landmarks and then converting them to images, which reduced the influence of non-audio related information in the video such as the angle of the shot. However, they only focus on the lip motion of the image. Song \emph{et al.} \cite{song2022everybody}, Thies \emph{et al.} \cite{thies2020neural} and Zhang \emph{et al.} \cite{zhang2021flow} all regressed facial expression parameters of 3DMM model, but they all relied on the video of the target portrait, which is not applicable in most scenarios. Furthermore, the 3DMM parameter can only represent the geometry of a face and does not render a natural talking head video with high-quality skin textures.

Based on the current state of research on talking head generation, we propose a novel two-stage talking head generation model, which guarantees the expressiveness of facial emotions while rendering the detailed skin texture of the target person well.

\vspace{-0.2cm}
\section{Method}

In this study, the model we proposed is shown in Figure~\ref{fig1}, which mainly includes two stages. The first stage is to predict the lip-sync and emotional face landmarks from the audio and the reference face image. The second stage is an attention-augmented translator based on U-net, which takes our predicted landmarks and the reference face image as input to generate photo-realistic and emotional talking head video frames. In the following subsections, each module is described in detail.

\subsection{Memory-Sharing Emotional Feature Extractor}
\begin{figure}[htbp]
\centerline{\includegraphics[width=0.9\linewidth]{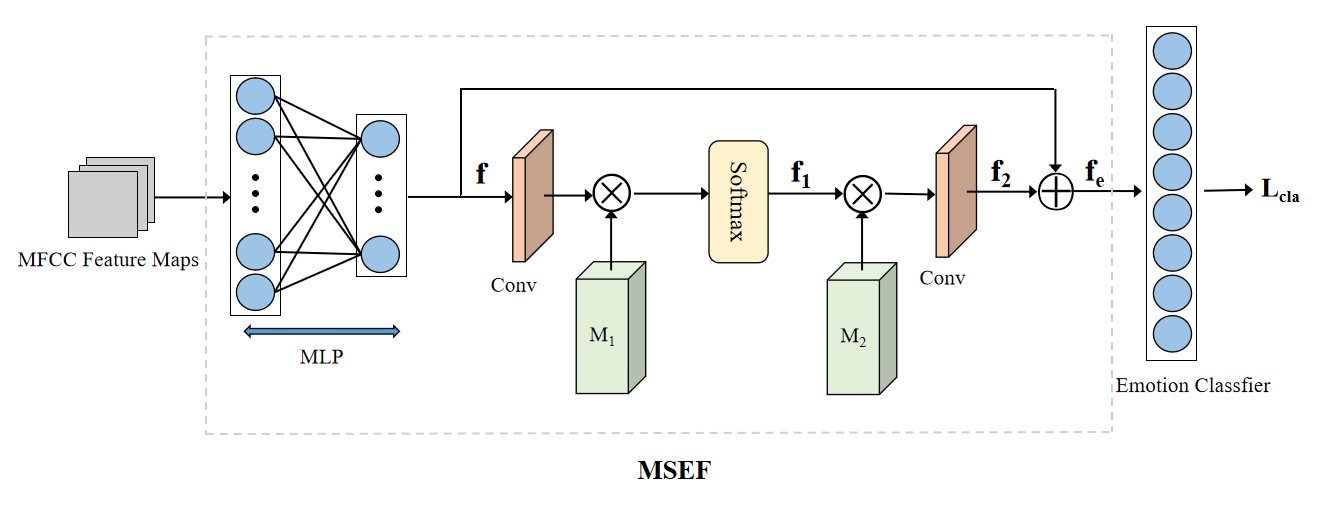}}
\caption{The structure of memory-sharing emotional feature extractor. $M_1$ and $M_2$ are two memory units. $L_{ec}$ is part of the joint loss.}
\label{msef}
\end{figure}

In order to more accurately predict the face landmarks of the avatar, we extract the implied emotion auxiliary features from the audio signal through the MSEF module, the specific network structure is shown in Figure~\ref{msef}.

We encode MFCC features as 128-dimensional feature vectors ($f$) through MLP, and introduce two memory units with reference to the idea of \cite{guo2022beyond}.  The memory units are represented by linear layers, which can easily capture the global features in a single sample and can focus on the potential correlation between different samples. The emotional information in an audio clip is global features, and different audios may contain the same emotion. We believe that the memory-sharing units can explore the correlation between different samples and thus extract emotional features more accurately. Features that correspond to the same emotion with different audio, the treatment should be consistent. The specific formula is as follows:
\begin{equation}
\begin{split}
    f_e = f + g(softmax(g(f) \cdot M_1) \cdot M_2),
\label{equ:memory}  
\end{split}
\end{equation}

\noindent where g($\cdot$) represents the convolution operation, and the rest symbols are shown in Figure~\ref{msef}.

Then, we encode the emotional features into  8-dimensional feature vectors via an additional emotion classifier and introduce $L_{ec}$ to supervise the training of this module. The loss function is formulated as follows:
\begin{equation}
\begin{split}
    L_{ec} = \frac{1}{N} \sum_{i=1}^{N} - [ y_i ln \hat{y}_{i} + (1 - y_i) ln(1 - \hat{y}_{i})],
\label{equ:ec}  
\end{split}
\end{equation}
\noindent
where $y$ represents the real emotion label of audio and $\hat{y}$ represents the predicted emotion category.

In addition, $E_{Lm}$ and $E_A$ are simple MLPs, encoding the landmark and MFCC into 512-dimension and 128-dimension feature vectors respectively. The $Audio2Lm$ module is composed of LSTM and a full connection layer. 

In order to consider emotion without losing the accuracy of lips, we designed a joint loss function. In addition to the $L_{ec}$ mentioned above, we add $L_{landmark}$ to the loss function to give the model the ability to regress face landmarks. At the same time, $L_{lip}$ is added to make the model pay more attention to lips. The specific formulas are as follows:
\vspace{-0.2cm}

\begin{equation}
\begin{split}
    L_{landmark} = \frac{1}{N} \sum_{i=1}^{N} (L_{real} - L_{fake})^{2},
\label{equ:landmark}  
\end{split}
\end{equation}
\begin{equation}
\begin{split}
    L_{joint} = L_{pca} + \alpha L_{landmark} + \beta L_{lip} + \gamma L_{ec}.
\label{equ:joint}  
\end{split}
\end{equation}

\noindent
where the hyperparameter $\alpha$, $\beta$ and $\gamma$ are the scaling factors, which we set to 10. $L_{real}$ is the real face landmark and $L_{fake}$ is the predicted face landmark. $M_{real}$ is the real lip landmark, $M_{fake}$ is the predicted lip landmark. $L_{pca}$ and $L_{lip}$ are calculated in a similar way to $L_{landmark}$, denoting the pca downscaling landmark and landmark of the lip region, respectively.

\subsection{Attention-Augmented Translator based on U-net}

In order to generate high-fidelity and emotional talking head video frames of the target person from the predicted landmarks, two challenges will be faced. Firstly, the photo-realistic talking head video frames need to pay attention to skin texture, and other details in order to better express emotions. Secondly, during the conversion from face landmarks to talking head video frames, a high degree of consistency with the target person's identity and a match to the predicted landmark facial contours and lip shape needs to be ensured.

\begin{figure}[htbp]
\centerline{\includegraphics[width=0.9\linewidth]{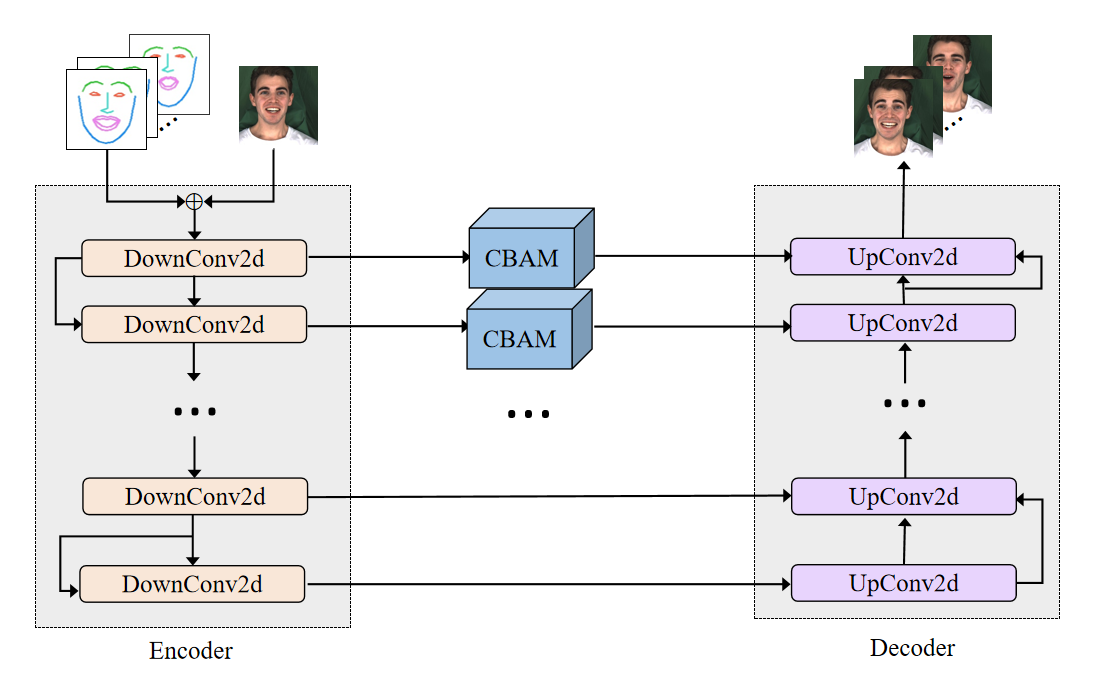}}
\caption{The structure of attention-augmented translator based on U-net. }
\label{aatu}
\end{figure}

In order to meet such challenges, we propose an attention-augmented translator based on U-net based on the MakeItTalk \cite{zhou2020makelttalk} framework. We propose AATU on this basis to further improve the quality of the generated video frames. As shown in Figure~\ref{aatu}, we concatenate the predicted face landmarks with the reference face image by channel, and take it as the input of the encoder. The output of the decoder is photo-realistic and lip-sync talking head video frames. In the initial four layers of the encoder and decoder, we add a CBAM \cite{woo2018cbam} module, respectively. 

The CBAM consists of two sub-modules, spatial attention and channel attention, and implements a sequential attention structure from channel to space. We believe that in this task, spatial attention enables the neural network to pay more attention to the pixel areas in the image that play a determining role in facial expression and lip shape, while ignoring the unimportant areas. Channel attention is used to handle the distribution relationship of the feature map channels. Also, the distribution of attention over the two dimensions reinforced the impact of the attention mechanism on model performance. The shallow layer of the U-net structure effectively avoids the loss of spatial information caused by the fully connected layer, allowing the network to pay attention to skin texture and other details.

We use L1 loss as the loss function to supervise and train our network, and in order to enhance the quality of the generated talking head video frames, additional perceptual loss \cite{johnson2016perceptual} is added.  The specific formula are as follows:
\begin{equation}
\begin{split}
    L1 = \frac{1}{N} \sum_{i=1}^{N} ||f - \hat{f}||,
\label{equ:l1}  
\end{split}
\end{equation}
\begin{equation}
\vspace{-0.2cm}
\begin{split}
    L_{per} = \frac{1}{N} \sum_{i=1}^{N} ||\phi_i(I) - \phi_i(\hat{I})||,
\label{equ:lper}  
\end{split}
\end{equation}
\noindent
 where $I$ represents the real image, $\hat{I}$ represents the generated image. $\phi_i$ represents the layer $i$ feature extraction layer of VGG-19 network \cite{simonyan2014very}.
\vspace{-0.2cm}
\section{Experiment And Result}

\subsection{Implement Details}

 \emph{1) Dateset and Setup.} The dataset we use to evaluate the model is the same as EVP \cite{ji2021audio}, \emph{i.e.,} the MEAD dataset \cite{wang2020mead}. Other datasets, such as LRW \cite{yang2019lrw} and VoxCeleb \cite{nagrani2017voxceleb}, are not suitable in our case, since they lack sentiment labels. And the CREMA-D \cite{cao2014crema} dataset does not distinguish much between various types of emotions. MEAD is a large-scale, high-quality emotional audio-visual dataset, which consists of 60 actors, including 8 basic emotions and 3 different emotional-intensity talking head videos. The training-test set is divided into a ratio of 8:2. We convert all talking head videos to 25fps and set the audio sample rate to 16KHz. For video                                  6  streams, we use Dlib to detect the face landmark of each frame. For audio streams, We extract MFCC at the window size of 25ms and hop size of 10ms. Our network is implemented using PyTorch. We use Adam optimizer, and the initial learning rate is set to 1e-4. We use the annealing strategy to adjust the learning rate through exponential decay. 

\emph{2) Evaluation Metrics.} In order to quantitatively evaluate different methods, we select common metrics in talking head generation. We used M-LMD and F-LMD to measure the accuracy of lip movements and facial contours. In addition, we use Structural Similarity Index Measure (SSIM) \cite{wang2004image} and Peak Signal to Noise Ratio (PSNR) \cite{narvekar2009no} to measure the quality of the generated talking head video frames.

\emph{3) Compared Methods.} To the best of our knowledge, there are now open-source works that consider emotional information, such as EVP \cite{ji2021audio} and EAMM \cite{ji2022eamm}. However, EAMM is speaker-independent when partitioning the dataset, and our approach is speaker-dependent in the same way as EVP. To be fair, we have compared our work with EVP, and our baseline model is based on ATVG \cite{chen2019hierarchical} and MakeItTalk \cite{zhou2020makelttalk}. In addition, we also have compared with Audio2Head \cite{wang2021audio2head} , which is based on motion fields to generate talking head videos, and improved the realism of videos from the perspective of generating head movements.

\vspace{-0.2cm}

\subsection{Quantitative Result}

\vspace{-0.2cm}
\begin{table}[htbp]
\caption{Quantitative results on the MEAD test set. $\uparrow$ means the higher the better, $\downarrow$ means the lower the better.}
\label{tab:eval}
\centering
\setlength{\tabcolsep}{1.5mm}
\begin{tabular}{c | c | c | c | c}
\hline
 \textbf{Method} & \textbf{F-LMD$\downarrow$} &  \textbf{M-LMD$\downarrow$} & \textbf{SSIM$\uparrow$} &  \textbf{PSNR$\uparrow$}\\
 \hline
Ground Truth & 0.00 & 0.00 & 1.00 & N/A\\
\hline
Baseline & 2.39 & 3.38 & 0.69 & 32.38 \\
\hline
EVP \cite{ji2021audio} & 3.01 &  \textbf{2.45} & 0.71 &  29.53\\
\hline
Audio2Head \cite{wang2021audio2head} & - & - & 0.69 & 30.91\\
\hline
Ours w/o MSEF & 2.38 & 3.06 & 0.72 & 33.30 \\
\hline
Ours w/o AATU & 2.35 & 3.02 & 0.72 & 33.29\\
\hline
Ours & \textbf{2.35} & 3.02  & \textbf{0.72} & \textbf{33.32}\\
\hline
\end{tabular}
\end{table}

\vspace{-0.4cm}

 ``Ours w/o MSEF" represents only add AATU model, and ``Ours w/o AATU" represents only add MSEF model. As can be seen from Table \ref{tab:eval}, when both MSEF and AATU are added, our model shows improvements in both emotion representation and image quality. Compared to EVP, our results show an increase of 0.66 in F-LMD and 3.79 in PSNR. The module proposed by EVP is helpful for lip accuracy. However, the module relies on long video drives of neutral emotions of the target person, which requires filtering audio pairs of the same content and different emotions, and is slightly weaker in terms of the intensity of emotional expression.
 
 Because Audio2Head is not a landmark-based method, we have only compared the latter two metrics with it. Again, our method outperforms Baseline and Audio2Head in all metrics. The baseline lacks emotional information as an auxiliary secondary feature and is slightly less effective in emotional face fitting. Audio2Head generates pixel-level talking head video frames based on motion fields, losing some important information about the speaker and resulting in limited image quality generated by their method.

\subsection{Qualitative Result}

\begin{figure}[htbp]
\centerline{\includegraphics[width=1.0\linewidth]{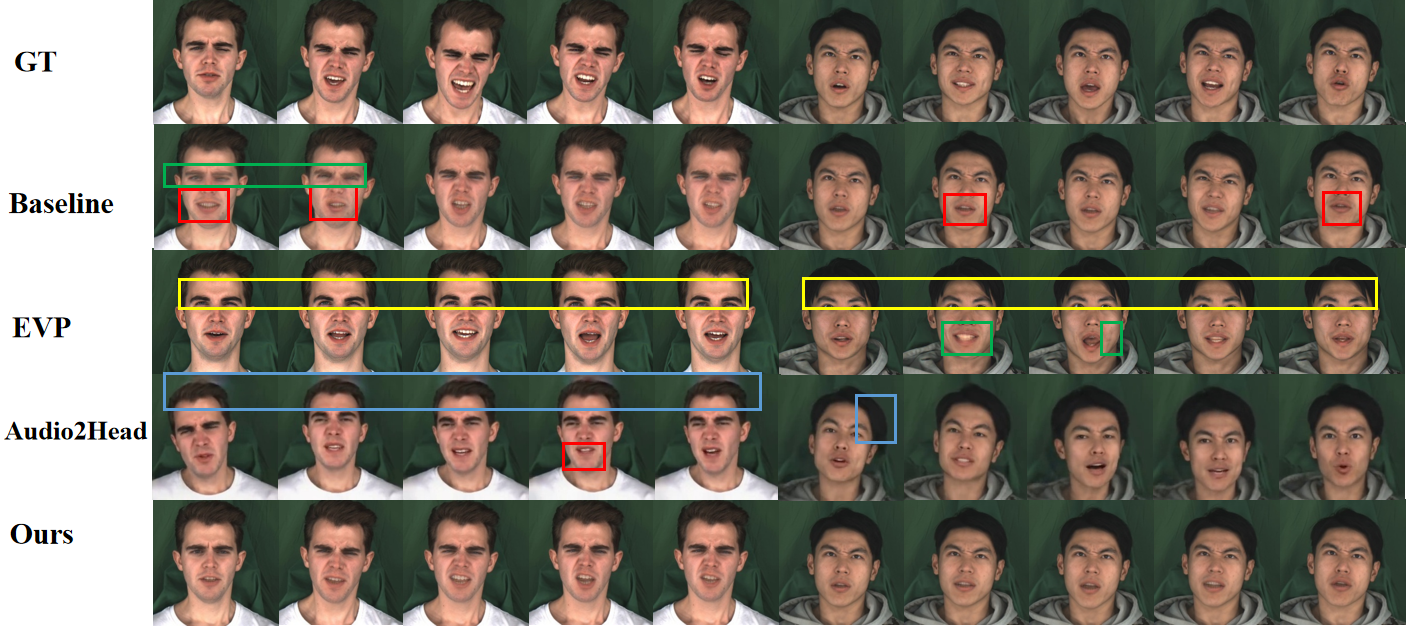}}
\caption{Qualitative results on the MEAD test set. }
\label{compare}
\end{figure}

\vspace{-0.2cm}

In order to visualize our comparison results, we also select some talking head video frames. As shown in Figure~\ref{compare}, our method generates high realistic talking head videos with strong emotions. Specifically, the yellow box locations are deficient in emotional expressiveness, the green box locations produce subtle artifacts, the red box locations have poor lip synchronization, and the blue box locations have poor identity consistency effects.

\begin{figure}[htbp]
\setlength{\abovecaptionskip}{0cm}
\centerline{\includegraphics[width=1.0\linewidth]{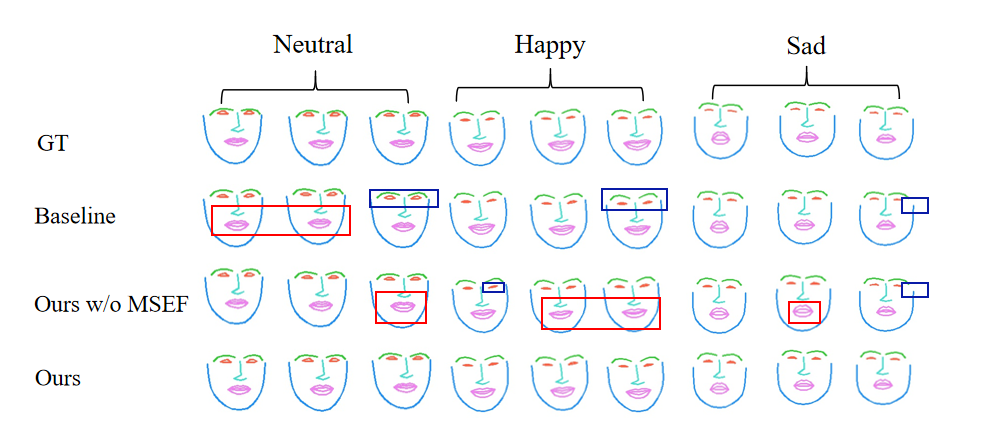}}
\caption{Landmarks generated by different methods. Lip synchronization is inaccurate for the position marked by the red box, while landmarks of facial expressions are inaccurate for the position marked by the blue box (better view by zooming in).}
\label{lmk}
\end{figure}

\vspace{-0.3cm}

To further see the contribution of MSEF module to the accuracy of landmarks regression, we visualized the landmarks generated by different methods. It can be seen from Figure~\ref{lmk} that the landmarks generated after adding the MSEF module are closest to the ground truth. Specifically, the red box locations are inaccurate concerning the lip shapes, and the blue box locations are inaccurate concerning eye shapes and facial contours.

\subsection{User Study}

\begin{table}[htbp]
\caption{User study results of LS, EE and VPQ.}
\label{tab:user}
\centering
\setlength{\tabcolsep}{4mm}
\begin{tabular}{c | c | c | c }
\hline
 \textbf{Method} & \textbf{LS$\uparrow$} &  \textbf{EE$\uparrow$} & \textbf{VPQ$\uparrow$} \\
\hline
Baseline & 4.32 & 6.97 & 5.03 \\
\hline
EVP \cite{ji2021audio} & \textbf{5.83} &  5.55 & 5.29\\
\hline
Audio2Head \cite{wang2021audio2head} & 5.07 & 5.90 & 5.59 \\
\hline
Ours & 5.25 & \textbf{7.49} & \textbf{5.82}\\
\hline
\end{tabular}
\end{table}

\vspace{-0.3cm}

In addition, we designed a detailed user study to assess the overall quality of the talking head videos. We used three metrics to measure video quality, \emph{i.e.}, \emph{Lip Synchronization} (LS), \emph{Emotional Expressiveness} (EE) and \emph{Video-Perceived Quality} (VPQ). A total of 30 participants completed our experimental questionnaire and they were asked to rate each video in the questionnaire from 1 (worst) to 10 (best). As Table~\ref{tab:user} shows, although our lip sync is slightly worse than EVP, it is superior in terms of emotional expressiveness and video perception quality. This is because, EVP mainly focuses on lip synchronization, and our method works on whole facial modeling. Moreover, our method outperforms Baseline and Audio2Head on all metrics.

\vspace{-0.2cm}

\section{Conclusion}
In this work, we propose a novel emotional talking head generation model, which consisted of a memory-sharing emotional feature extractor and an attention-augmented translator based on U-net. MSEF module is proposed to better predict the face landmarks in the talking head video. AATU module is proposed to better fit the facial details in the frames and improve the talking head video perception quality. Extensive experiments have proved that our method can generate lip-sync and emotional talking head videos. In the future, we will consider adding personalized head movements to the videos to further enhance the realism.
\bibliographystyle{IEEEtran}
\bibliography{mybib}

\end{document}